\documentclass{article}
\usepackage{spconf,amsmath,graphicx,pdfpages,hyperref,multirow,color,amssymb}

\usepackage{amsmath,amsfonts,bm}

\def\eqref#1{equation~\ref{#1}}

\def\1{\bm{1}}

\def\vh{{\bm{h}}}

\def\vp{{\bm{p}}}

\def\mC{{\bm{C}}}

\def\mT{{\bm{T}}}

\DeclareMathAlphabet{\mathsfit}{\encodingdefault}{\sfdefault}{m}{sl}
\SetMathAlphabet{\mathsfit}{bold}{\encodingdefault}{\sfdefault}{bx}{n}

\def\gC{{\mathcal{C}}}

\def\gL{{\mathcal{L}}}

\def\gT{{\mathcal{T}}}

\newcommand{\E}{\mathbb{E}}

\newcommand{\softmax}{\mathrm{softmax}}

\DeclareMathOperator*{\transformerenc}{Trans-Enc}
\DeclareMathOperator*{\mlp}{MLP_{clf}}
\DeclareMathOperator*{\mlpc}{MLP_{cont}}
\DeclareMathOperator*{\mlpt}{MLP_{type}}
\DeclareMathOperator*{\mlpd}{MLP_{dis}}
\DeclareMathOperator*{\sig}{Sigmoid}

\title{Disentangled and Robust Representation Learning \\
for Bragging Classification in Social Media}

\name{Xiang Li$^1$, Yucheng Zhou$^2$}
\address{$^1$College of Intelligence and Computing, Tianjin University, \\ $^2$Australian AI Institute, School of Computer Science, FEIT, University of Technology Sydney\\
lixiang\_eren@tju.edu.cn, yucheng.zhou-1@student.uts.edu.au}

\begin{document}
\maketitle
\begin{abstract}
Researching bragging behavior on social media arouses interest of computational (socio) linguists. However, existing bragging classification datasets suffer from a serious data imbalance issue. Because labeling a data-balance dataset is expensive, most methods introduce external knowledge to improve model learning. Nevertheless, such methods inevitably introduce noise and non-relevance information from external knowledge. To overcome the drawback, we propose a novel bragging classification method with disentangle-based representation augmentation and domain-aware adversarial strategy. Specifically, model learns to disentangle and reconstruct representation and generate augmented features via disentangle-based representation augmentation. Moreover, domain-aware adversarial strategy aims to constrain domain of augmented features to improve their robustness. Experimental results demonstrate that our method achieves state-of-the-art performance compared to other methods.
\end{abstract}


\section{Introduction}
\label{sec:intro}
Bragging classification aims to predict the bragging type for a social media text. As online communication on social media is more pervasive and essential in human life, bragging (or self-promotion) classification has become a significant area in computational (socio) linguistics \cite{Wang2021SelfPI,JinPDA22}. It has been widely applied in academia and industry, like helping linguists dive into the context and types of bragging \cite{JinPDA22}, supporting social scientists to study the relation between bragging and other traits (e.g., gender, age, economic status, occupation) \cite{Wang2021SelfPI,HowAnonymityInfluence}, enhancing online users' self-presentation strategies \cite{ManagingImpressions,VanDamme2017WhenPY}, and many real-world NLP applications in business, economics and education \cite{Prinsloo2021ResponsesTB,Kerr2012BraggingRA}.

Although bragging has been widely studied in the context of online communication and forum, all these studies depend on manual analyses on small data sets  \cite{EmotionalConsequences,ManagingImpressions,ManbraggingOnLine,TellingTheWorld,HowAnonymityInfluence}. To efficiently research bragging on social media, Jin \textit{et al.} \cite{JinPDA22} collect the first large-scale dataset of bragging classification in computational linguistics, which contains six bragging types and a non-bragging type. However, the dataset suffers from a heavy data imbalance issue. For example, there are 2,838 examples in the non-bragging type, while only 58 to 166 (i.e., 1\% $\sim$ 4\%) in the other bragging types. It severely affects the learning of the model on examples with these bragging types.

To alleviate the data imbalance issue, apart from employing a weighted loss function to balance sample learning from different types \cite{dice,focalloss}, many researchers attempt to perform data augmentation by injecting models with external knowledge, such as knowledge graph \cite{Zhou21Modeling,Yu2022JAKETJP}, pre-trained word embedding \cite{w2v,glove}, translation \cite{Zhou21Improving} and some related pragmatics tasks, i.e., complaint severity classification \cite{Filgueiras2019ComplaintAA}. As for bragging classification, Jin \textit{et al.} \cite{JinPDA22} inject language models with external knowledge from the NRC word-emotion lexicon, Linguistic Inquiry and Word Count(LIWC) and vectors clustered by Word2Vector. Despite their success, improvement of external knowledge injection relies on the relevance between bragging classification and other pragmatics tasks. However, knowledge provided by other pragmatic tasks is fixed and obtained in a model-based manner, which inevitably brings noise.

To get rid of the noise from external knowledge injection, we propose a disentangle-based feature augmentation for disentangled representation and augmented feature learning without any other external knowledge. Specifically, we first disentangle content and bragging-type information from a representation. Next, we generate a reconstructed representation by integrating disentangled information and then constrain consistency between representation and reconstructed representation. To address the data imbalance problem, we fuse disentangled information from different examples to generate augmented features for model training.

\begin{figure*}[t]
  \centering
  \includegraphics[width=0.9\linewidth]{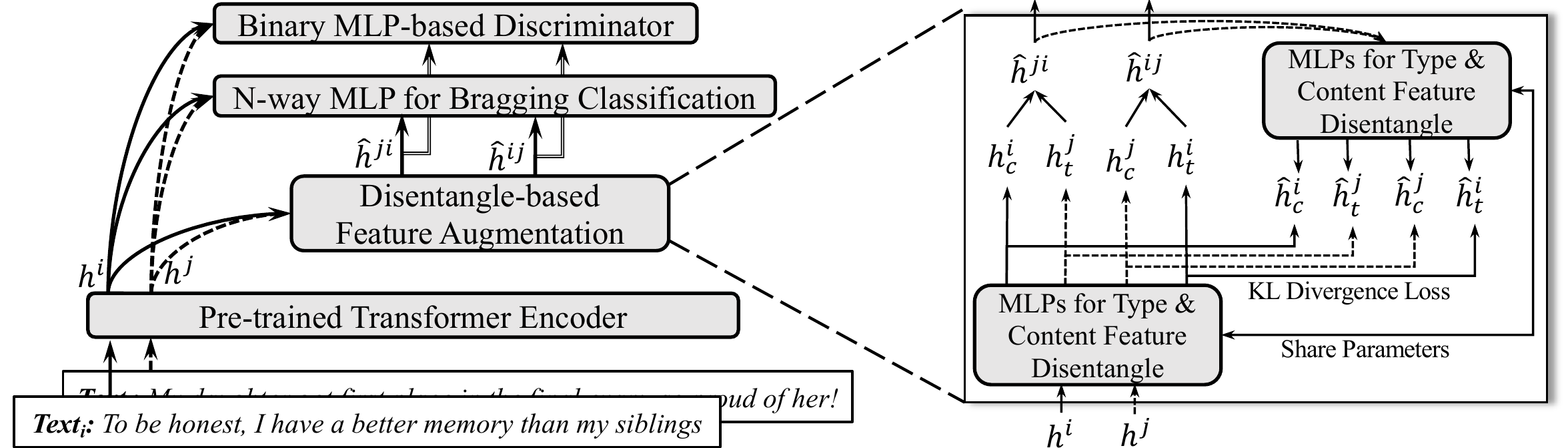}
  \caption{Overview of our model.}
  \label{fig:model}
\end{figure*}

Moreover, we propose a domain-aware adversarial strategy to mitigate domain disorder caused by augmented features. Specifically, we present a discriminator on top of the language model, which is trained to distinguish whether the input is a representation from the encoder or an augmented feature. Meanwhile, jointing with a classification objective, the encoder is trained to fool the discriminator, which pushes the model to generate robust augmented features that are domain consistent with representations from the encoder.

In the experiments, we train and evaluate our method on bragging classification dataset \cite{JinPDA22}. The results show that our method achieves state-of-the-art performance compared to other strong competitors. 

\section{Method}
\label{sec:method}
This section starts with a base bragging classification model, followed by our proposed methods, i.e., disentangle-based feature augmentation and domain-aware adversarial strategy. Lastly, training details are elaborated.

\subsection{Base Bragging Classification Model}
The bragging classification task aims to automatically classify bragging type of a given text from social media. Since pre-trained language models show their excellent performance in natural language processing (NLP) tasks, the general paradigm for NLP classification tasks is to fine-tune pre-trained language models (e.g. BERT \cite{DevlinCLT19}, RoBERTa \cite{Liu2019RoBERTaAR}). In this work, our base bragging classification model is composed of a pre-trained transformer encoder and an MLP with $\softmax$. Given a text $\mT_i$, the model is to distinguish its bragging type $c$, i.e, 
\begin{align}
    \vh^i &= \transformerenc(\mT_i;\theta^{(enc)}) \label{equ:enc} \\
    \vp^i &= \softmax(\mlp(\vh^i;\theta^{(clf)})) \label{equ:clf}
\end{align}
where $i$ denotes $i$-th training example; $\vh^i$ denotes representation of the text $\mT_i$, and $\vp^i$ refers to a probability distribution over all bragging types $\mC$.
Lastly, we train the pre-trained transformer encoder $\theta^{(enc)}$ and MLP $\theta^{(clf)}$ by maximum likelihood estimation, and its loss function is written as:
\begin{align}
    \gL^{(clf)} = - \dfrac{1}{|\gT|}\sum_{\gT}\log\vp^i_{[y=c]}, \label{equ:loss_cls}
\end{align}
where $\gT$ denotes a collection of training texts; $c$ refers to a ground truth bragging type, and $c \in \gC$.

\subsection{Disentangle-based Feature Augmentation}
To alleviate data imbalance problem, previous works \cite{JinPDA22} propose data augmentation to enrich text embedding representation by injecting external knowledge from the NRC word-emotion lexicon, Linguistic Inquiry and Word Count(LIWC) and vectors clustered by Word2Vector. However, these methods achieve only a limited gain (i.e., -0.5\% $\sim$ 1.09\% on F1) compared to the base bragging classification model. The reason is that injected external knowledge is fixed (not trainable) and is full of noise information unrelated to bragging classification, which interferes with the model training. Therefore, we introduce a disentangled-based feature augmentation method instead of external knowledge injection. 

Since the text is encoded as representation,  it is reasonable to assume that representations contain separable content and specific features from a feature disentanglement perspective \cite{Bengio2013RepresentationLA}. Therefore, we separate the representation $\vh^i$ into two features: one is closely related to bragging types, denoted as $\vh^i_t$, and the other contains separable content unrelated to bragging, denoted as $\vh^i_c$. To disentangle these two features from representation $\vh^i$, we employ two MLPs for content and bragging type features disentangle, respectively, i.e., 
\begin{align}
    \vh^i_c &= \mlpc(\vh^i;\theta^{(cont)})  \label{equ:mlpc}\\
    \vh^i_t &= \mlpt(\vh^i;\theta^{(type)})  \label{equ:mlpt}
\end{align}
where $\vh^i_c$ and $\vh^i_t$ denote content and bragging type disentangled features, respectively.

Based on disentangled features, we can obtain new augmented features by integrating content and bragging type features from different samples, i.e,
\begin{align}
    \hat{\vh}^{ji} = \vh^i_c + \vh^j_t
\end{align}
where $\hat{\vh}^{ji}$ denotes an augmented feature; $\vh^j_t$ refers to a bragging type feature from $j$-th training example. It is remarkable that bragging type of $\hat{\vh}^{ji}$ is the same as bragging type feature $\vh^j_t$ because it contains bragging type features $\vh^j_t$. Augmented feature $\hat{\vh}^{ji}$ is passed into Eq.\ref{equ:clf} to derive a probability distribution $\vp^{ji}$, and loss function can be defined as:
\begin{align}
    \gL^{(clf^+)} = - \dfrac{1}{|\gT|}\sum_{\gT}\log\vp^{ji}_{[y=c]}, \label{equ:loss_cls}
\end{align}

Moreover, we employ Kullback-Leibler (KL) divergence loss to guide
representation disentangling and representation reconstruction. Specifically, we disentangle $\hat{\vh}^{ji}$ to reconstruct content feature $\hat{\vh}^i_c$ and bragging type feature $\hat{\vh}^j_t$ by Eq.\ref{equ:mlpc} and Eq.\ref{equ:mlpt}. Then, we optimize the reconstruction loss by minimizing KL divergence, defined as follow:
\begin{align}
    \notag \gL^{(kl)} &= \E(\log \vh^x_y - \log \hat{\vh}^z_y), \\
    &\text{where,}~ x,z \in \gT, y \in \{c,t\} \label{equ:loss_kl}
\end{align}

\subsection{Domain-aware Adversarial Strategy}
Although model trained on augmented features circumvents data imbalance problem, it inevitably suffers from domain discrepancy between encoder representations and augmented features, which is verified to undermine the performance \cite{Hong2021DomainAwareUS}. We thus design an adversarial strategy to constrain the augmented features to follow the same domain with representations. Formally, a discriminator is built upon representation and augmented features:
\begin{align}
    \notag p^{(adv)} &= \sig(\mlpd(\vh; \theta^{(dis)})), \\
    &\text{where,}~ \vh \in \{\vh_i, \vh_j, \hat{\vh}^{ji}, \hat{\vh}^{ij}\}.
\end{align}
$p^{(adv)}$ is the probability of the feature not augmented. The discriminator is trained to minimize:
\begin{align}
\gL^{(dis)} = -\mathbb{I}_{(rep)} \log \vp^{(adv)} - \mathbb{I}_{(aug)} \log(1-\vp^{(adv)}) \label{equ:loss_dis}
\end{align}
where $\mathbb{I}_{(aug)}$ denotes if the feature is an augmented feature; $\mathbb{I}_{(rep)}$ denotes if the feature is encoder representation. On the contrary, augmented features are learned to fool by minimizing an adversarial loss, i.e., 
\begin{align}
 \gL^{(adv)} = - \mathbb{I}_{(aug)} \log \vp^{(adv)}. \label{equ:loss_gen}
\end{align}

\subsection{Training}
During training, we alternately train model and discriminator. we train our model jointly by minimizing four loss functions:
\begin{align}
 \gL\!\!=\!\!\alpha \times \gL^{(clf)} \!+\! \beta \times \gL^{(clf^+)} \!+\! \lambda \times \gL^{(kl)} \!+\! \gamma \times \gL^{(adv)} \label{equ:loss}
\end{align}

Meantime, discriminator is trained by $\gL^{(dis)}$.

\section{Experiments}
\label{sec:exp}

\subsection{Dataset and Evaluation Metrics}
In experiments, we train and evaluate our proposed approach on the dataset built by \cite{JinPDA22}. This dataset is collected from tweeter by Premium Twitter Search API and further annotated by human-craft. Bragging types include ``Not Bragging", ``Achievement", ``Action", ``Feeling", ``Trait", ``Possession" and ``Affiliation". We follow the official data split with 3,382/3,314 samples in training/test sets. Following \cite{JinPDA22}, evaluation metrics are macro precision, recall and F1 score.

\subsection{Experimental Setting}
The pre-trained transformer encoder we used is BERTweet \cite{nguyen-etal-2020-bertweet}. We use AdamW optimizer\cite{DBLP:journals/corr/KingmaB14,DBLP:conf/iclr/LoshchilovH19} with learning rate of $3 \times 10^{-6}$ for training, and learning rate for discriminator is $3 \times 10^{-4}$. The maximum training epoch and batch size are set to 26,500 and 35. The maximum sequence length, weight decay and gradient clipping are set to 128, 0.01 and 1.0. The dropout of model and discriminator are set to 0.2 and 0.5, separately.
In Eq.\ref{equ:loss}, $\alpha$, $\beta$, $\lambda$ and $\gamma$ are set to 1.5, 1.5, 1.0 and 1.5, respectively. Experiments are conducted on a NVIDIA RTX2070 GPU, and training time is around 5 hours.

\subsection{Main Results}
\begin{table}[t]\small
\centering
\begin{tabular}{lccc}
\hline
\multirow{2}{*}{\textbf{Method}} & \multicolumn{3}{c}{\textbf{Macro Average}}                                                                            \\ \cline{2-4} 
                                 & \multicolumn{1}{c}{\textbf{Precision}} & \multicolumn{1}{c}{\textbf{Recall}} & \multicolumn{1}{c}{\textbf{F1 Score}} \\ \hline
Majority Class \cite{JinPDA22}                  & 13.26                                  & 14.29                               & 13.76                                  \\
LR-BOW \cite{JinPDA22}                          & 18.52                                  & 20.02                               & 18.59                                  \\
BiGRU-Att \cite{JinPDA22}                        & 18.32                                  & 26.16                               & 19.19                                  \\\hline
BERT \cite{DevlinCLT19}                             & 24.16                                  & 39.66                               & 26.85                                  \\
RoBERTa \cite{Liu2019RoBERTaAR}                          & 28.99                                  & 45.90                               & 32.82                                  \\
BERTweet \cite{nguyen-etal-2020-bertweet}                         & 30.82                                  & 47.25                               & 34.86                                  \\
BERTweet-NRC \cite{JinPDA22}                    & 30.95                                  & \textbf{47.98}                      & 34.36                                  \\
BERTweet-LIWC \cite{JinPDA22}                   & 32.06                                  & 46.68                               & 34.83                                  \\
BERTweet-Clusters \cite{JinPDA22}               & 32.51                                  & 46.97                               & 35.59                                  \\
Ours                             & \textbf{41.18}                         & 40.08                               & \textbf{39.86}                         \\ \hline
\end{tabular}
    \caption{\small Bragging multi-classification results}
    \label{tab:main}
\end{table}
\begin{table}[t] \small
\centering
\begin{tabular}{lccl}
\hline
\multirow{2}{*}{\textbf{Method}} & \multicolumn{3}{c}{\textbf{Macro Average}}               \\ \cline{2-4} 
                        & \textbf{Precision} & \textbf{Recall} & \textbf{F1 Score} \\ \hline
Ours                    & 41.18     & 40.08  & \textbf{39.86}    \\
Ours w/o DAS             & \textbf{45.07}              & 35.02           & 38.84 (-1.02)     \\
Ours w/o DAS, DFA        & 30.82              & \textbf{47.25}           & 34.86 (-3.98)     \\ \hline
\end{tabular}
\caption{\small Ablation study of our approach. 
``\textit{w/o DAS}'' denotes removing the domain-aware adversarial strategy in our model, 
and ``\textit{w/o DAS, DFA}'' indicates removing both disentangle-based feature augmentation and domain-aware adversarial strategy.}
\label{tab:ablation}
\end{table}
Comparison results are shown in Table \ref{tab:main}. On the table, it is clear that 1) using pre-trained language model as base model can significantly improve performance, mainly because pre-trained language models have stronger text representation capability; 2) our approach is superior to the method proposed by \cite{JinPDA22} and achieves state-of-the-art,demonstrating the proposed disentangled and robust representation learning is effective. From these two observations above, it is insightful to assert that learning better representation from data itself is more critical than external knowledge injection.
\begin{table}[t]\small
\centering
\begin{tabular}{lccc}
\hline
\multirow{2}{*}{\textbf{Method}} & \multicolumn{3}{c}{\textbf{Macro Average}}                                                                            \\ \cline{2-4} 
                                 & \multicolumn{1}{c}{\textbf{Precision}} & \multicolumn{1}{c}{\textbf{Recall}} & \multicolumn{1}{c}{\textbf{F1 Score}} \\ \hline
BERT \cite{DevlinCLT19}                          & 64.24                                  & 65.91                               & 64.58                                  \\
RoBERTa \cite{Liu2019RoBERTaAR}                  & 66.53                                  & 68.43                               & 67.34                                  \\
BERTweet \cite{nguyen-etal-2020-bertweet}        & 70.43                                  & \textbf{72.62}                      & 71.44                                  \\
BERTweet-NRC \cite{JinPDA22}                     & 72.89                                  & 70.95                               & 71.80                                  \\
BERTweet-LIWC \cite{JinPDA22}                    & 72.65                                  & 72.21                               & 72.42                                  \\
BERTweet-Clusters \cite{JinPDA22}                & 71.26                                  & 72.53                               & 71.60                                  \\
Ours                             & \textbf{78.42}                         & 69.67                               & \textbf{73.07}                         \\ \hline
\end{tabular}
    \caption{\small Bragging binary-classification results.}
    \label{tab:bclf}
\end{table}
\begin{table}[t] \small
\centering
\begin{tabular}{lccc}
\hline
\multirow{2}{*}{\textbf{Method}} & \multicolumn{3}{c}{\textbf{Macro Average}}                                                                           \\ \cline{2-4} 
                        & \textbf{Precision}                  & \textbf{Recall}                  & \textbf{F1 Score}                  \\ \hline
BERTweet \cite{nguyen-etal-2020-bertweet}               & 30.82                               & \textbf{47.25}                            & 34.86                              \\
Only $\vh^i_t$                   & \textbf{43.94}                               & 31.00                            & 33.96                              \\ 
Ours                    & 41.18                               & 40.08                           & \textbf{39.86}                              \\\hline
\end{tabular}
    \caption{\small Impact of disentangled type feature in our method. ``\textit{Only $\vh^i_t$}'' denotes classification using bragging type disentangled features $\vh^i_t$ extracted by our model.}
    \label{tab:h_t}
\end{table}
\begin{figure}[t]
  \centering
  \includegraphics[width=0.9\linewidth]{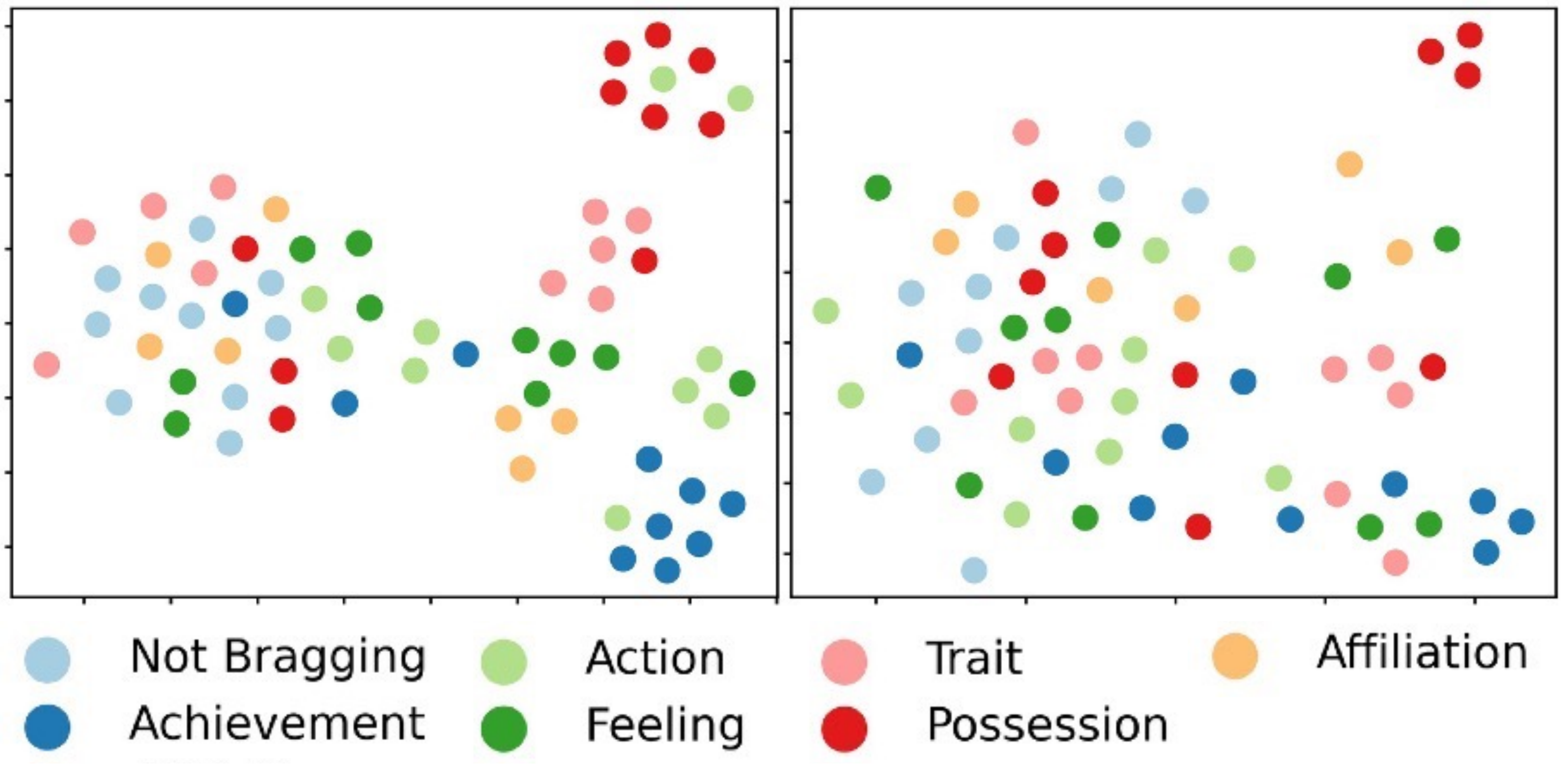}
  \caption{\small t-SNE visualization: ours (left) and BERTweet (right).}
  \label{fig:tsne}
\end{figure}
\begin{figure}[t]
  \centering
  \includegraphics[width=\linewidth]{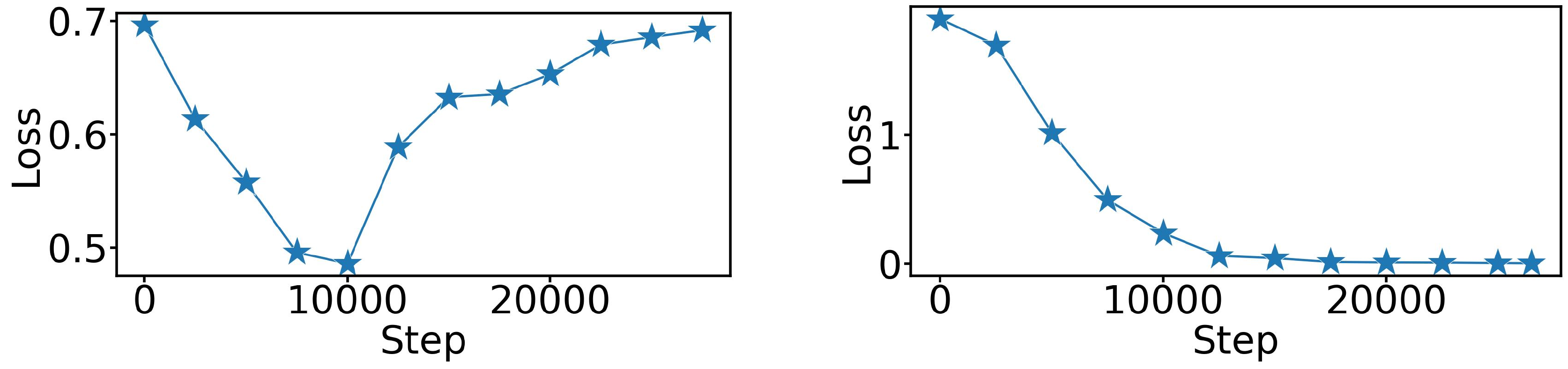}
  \caption{\small Classification loss (left) and discriminator loss (right) in training phase.}
  \label{fig:loss}
\end{figure}

\subsection{Ablation Study}
As shown in Table~\ref{tab:ablation}, we conduct an ablation study on our method. First, we remove domain-aware adversarial strategy, and results show that performance drops. It demonstrates that constraining the domain of augmented features using adversarial strategy is beneficial for model to learn robust representation. Moreover, we verify the effectiveness of augmented features by removing disentangle-based feature augmentation. Results show that the performance drops a lot, which demonstrates that disentangled representation learning is able to mine text information deeper.

\subsection{Binary Classification for Bragging}
To comprehensively evaluate our method, we also evaluate our model on bragging binary-classification task (i.e., bragging and non-bragging). As shown in Table \ref{tab:bclf}, we can observe that our model still achieves better performance as compared with other methods. Since data imbalance is greatly alleviated in binary-classification setting, the improvement of our method is not as significant as the multi-classification setting.

\subsection{Impact of Disentangled Type Feature}
To delve into the effectiveness of disentangled features, we employ MLP to classify disentangled feature $\vh^i_t$ directly, and results are shown in Table \ref{tab:h_t}. From Precision results, we can observe that Only $\vh^i_t$ outperforms BERTweet, which shows the effectiveness of feature disentanglement. Moreover, Only $\vh^i_t$ underperforms our method on F1 Score, which indicates that content feature disentanglement and domain consistency benefit bragging classification.

\subsection{Visualization Analysis}
We apply t-SNE visualization(Fig \ref{fig:tsne}.) on features obtained by Eq.\ref{equ:enc} to visually demonstrate how our approach works. It is obvious that scatters of our method represent a more clustering trend than BERTweet, which indicates our method is able to learn a type-relevant representation.

\subsection{Deep Dive into Adversarial Learning}
To investigate the impact of domain-aware adversarial strategy, we show classification loss and discriminator loss in Figure \ref{fig:loss}. The discriminator loss quickly drops and then slowly goes up, indicating that the discriminator is functional at first and fooled later by domain-consistency features. Meanwhile, The classification loss decreases rapidly and remains very low in the following steps, showing that domain-aware adversarial strategy is well integrated within the whole process.

\section{Conclusion}
In this study, to address the challenge of data imbalance, we propose a novel augmentation method for learning disentangled and robust representations without other external knowledge. The method includes disentangle-based feature augmentation and domain-aware adversarial strategy. Experimental result shows that our method achieves state-of-the-art performance. Lastly, we conducted extensive analyses to verify the effectiveness of our method.

\bibliographystyle{IEEEbib}
\bibliography{ref}

\end{document}